\documentclass[a4paper]{article}
\RequirePackage[normalem]{ulem}
\RequirePackage{color}\definecolor{RED}{rgb}{1,0,0}\definecolor{BLUE}{rgb}{0,0,1}

\usepackage[utf8]{inputenc}
\usepackage{graphicx} 
\usepackage{authblk} 
\usepackage{amsmath} 
\usepackage{hyperref} 
\usepackage{graphicx, float}
\usepackage{textcomp} 
\usepackage{newunicodechar}
\usepackage{booktabs} 
\newunicodechar{≈}{\approx}
\usepackage{xcolor}

\usepackage[normalem]{ulem} 
\newunicodechar{Ω}{\textohm}

\usepackage{lineno}
\usepackage[a4paper,left=3cm,right=2cm,top=2.5cm,bottom=2.5cm]{geometry}
\newcounter{suppfigure}

\usepackage{xcolor}

\usepackage[isbn=false,maxbibnames=99,url=false,eprint=false,sorting=none,citestyle=numeric-comp]{biblatex}
\usepackage{subcaption}

\usepackage{multirow}

\title{Personalized Spiking Neural Networks with Ferroelectric Synapses for EEG Signal Processing}

\author[1]{Nikhil Garg}
\author[1]{Anxiong Song}
\author[1]{Niklas Plessnig}
\author[1]{Nathan Savoia}

\author[1]{Laura Bégon-Lours}

 {}

\affil[1]{Integrated Systems Laboratory, Department of Information Technology and Electrical Engineering, ETH Zürich, CH-8092 Zürich, Switzerland
\newline
Email Address: nigarg@ethz.ch, lbegon@ethz.ch
}
\date{}

\begin{document}
\maketitle

\textit{Cite as}: Nikhil Garg, Anxiong Song, Niklas Plessnig, Nathan Savoia and Laura Bégon-Lours, \textit{Personalized Spiking Neural Networks with Ferroelectric
Synapses for EEG Signal Processing}, APL Machine Learning (2026), DOI: 10.1063/5.0319912 \\

{\noindent\textbf{Abstract}

Electroencephalography (EEG)-based brain-computer interfaces (BCIs) are strongly affected by non-stationary neural signals that vary across sessions and individuals, limiting the generalization of subject-agnostic models and motivating adaptive and personalized learning on resource-constrained platforms. Programmable memristive hardware offers a promising substrate for such post-deployment adaptation; however, practical realization is challenged by limited weight resolution, device variability, nonlinear programming dynamics, and finite device endurance. In this work, we show that spiking neural networks (SNNs) can be deployed on ferroelectric memristive synaptic devices for adaptive EEG-based motor imagery decoding under realistic device constraints, achieving classification performance comparable to software-based SNNs. We fabricate, characterize, and model the weight update in ferroelectric synapses. We then evaluate the deployment of convolutional-recurrent SNN architecture using two strategies. First, we adapt to SNNs a mixed precision strategy in which gradient-based updates are accumulated digitally and converted into discrete programming events only when a threshold is exceeded. Additionally, the weight update is device-aware and accounts for the nonlinear, state-dependent programming dynamics. During learning and adaptation, this scheme mitigates possible endurance and energy constraints. Second, we evaluate the transfer of software-trained weights followed by low-overhead on-device re-tuning. We show that, subject-specific transfer learning achieved by retraining only the final network layers improves classification accuracy. These results demonstrate that programmable ferroelectric hardware can support robust, low-overhead adaptation in spiking neural networks, opening a practical path toward personalized neuromorphic processing of neural signals.


\section{Introduction}\label{sec1}

Brain-computer interfaces (BCIs) based on electroencephalography (EEG) have attracted significant interest due to their potential for assistive communication in patients with severe motor impairments \cite{bamdad2015application, kawala2021summary, kanungo2021wheelchair}. From a deployment perspective, practical EEG decoding must operate under stringent constraints on latency and power: decisions should be produced with minimal delay after the user’s intent while maintaining ultra-low energy consumption to ensure acceptable battery lifetime and to limit tissue heating in wearable and implantable settings. More broadly, wearable sensors and implantable devices are increasingly pushing artificial intelligence workloads toward the extreme edge \cite{wang2017flexible, shi2016edge, rancea2024edge}, where strict thermal and energy budgets fundamentally shape algorithm and hardware choices \cite{vazquez2021compute}. In this context, EEG-based decoding represents a representative extreme-edge signal-processing workload, requiring both low end-to-end latency and high energy efficiency.

A key challenge in extreme-edge biosignal processing is that conventional frame-based pipelines rely on continuous sampling, digitization, and dense processing, which increase energy consumption in sensing, data transmission, and downstream computation as latency requirements become tighter. In contrast, biological neural systems communicate through sparse, asynchronous events \cite{tayarani2021event}, where information is encoded in spike timing and temporal structure, offering an efficient representation for time-series biosignals \cite{garg2021signals}. This event-based sensing and processing paradigm motivates spiking neural networks (SNNs) \cite{maass1997networks}, which naturally capture temporal dynamics and provide intrinsic memory through neuronal and synaptic state. Beyond algorithms, brain-inspired computation also motivates architectures that co-localize memory and computation to reduce data movement, an overhead that dominates energy consumption in conventional CMOS processors \cite{horowitz20141}. Neuromorphic engineering seeks to translate these principles into event-driven hardware \cite{mead2020we, burr2017neuromorphic}, while in-memory computing aims to reduce data-transfer costs by blurring the boundary between storage and processing \cite{sebastian2020memory}. 

Within this architectural framework, beyond-CMOS memories have emerged as promising building blocks for in-memory computing \cite{theis2017end, mehonic2024roadmap}. In resistive memory arrays, physical laws such as Ohm’s and Kirchhoff’s laws can be exploited to perform multiply-and-accumulate operations directly within memory, reducing data transfer and its associated energy overhead. Memristive devices can emulate synaptic functionality by encoding network weights as programmable conductance states connecting successive layers. Their nonvolatile retention and sub-femtojoule read energy \cite{covi2022challenges}, together with demonstrated multi-level programmability \cite{thomas2024versatile}, make them attractive for adaptive processing of non-stationary physiological signals \cite{fang2022compact}. Analog or multi-state behavior suitable for learning has been demonstrated across material systems, including phase-change, valence-change memories \cite{rao2023thousands} and CMOS-compatible ferroelectric resistive weights \cite{begon2024back}.   Ferroelectric memristive synapses consist of an ultra-thin ferroelectric film (zirconium doped hafnium oxide) between two asymmetric electrodes. They are ideal candidates for neuromorphic hardware for four main reasons: programming is ultrafast, with low energy (allegedly below one nanosecond \cite{baigol2026analog} and one pJ \cite{begon2022effect}) and high endurance (above 1\textsuperscript{10} cycles \cite{begon2022scaled}).  In contrast, memristive technologies relying on ion motion or a change of phase typically require 10 pJ and can only endure 10\textsuperscript{6} cycles \cite{el2024toward}. Ferroelectric memristive weights can be read non-destructively at low voltages (\(< 100~\mathrm{mV}\)), with fJ energy \cite{begon2021high}. Finally, the ferroelectric switching dynamics mimic the biological, synaptic plasticity \cite{boyn2017learning}. Ferroelectric materials find applications beyond the field of neuromorphic hardware \cite{mikolajick2023ferroelectric}, for example as non-volatile memories \cite{silva2023roadmap}, photodetectors \cite{guo2025high}, energy harvesting devices \cite{ren2025ultrahigh}.

However, translating these motivations into practical EEG decoding systems remains challenging. Memristive synapses are constrained by limited effective weight resolution, device-to-device variability, nonlinear and state-dependent programming dynamics, and finite endurance, all of which can degrade accuracy and stability when deploying learned models on hardware. At the system level, two common deployment strategies are (i) offline learning in software followed by weight transfer through device programming, and (ii) direct on-device learning using local update rules \cite{garg2025unsupervised, alibart2013pattern}. Both strategies must contend with the highly non-stationary nature of EEG and the significant variability across trials, recording sessions, and individuals, which limits the generalization of subject-agnostic models and motivates adaptive and personalized learning. At the same time, personalization \cite{zheng2016personalizing} must be achieved under strict constraints on energy consumption, programming frequency, and device lifetime, requiring learning mechanisms that are robust to device non-idealities while enabling low-overhead post-deployment adaptation.

In this work, we investigate the deployment of SNNs for EEG-based motor imagery decoding using ferroelectric memristive synaptic devices. The chosen technology is CMOS compatible: in ref. \cite{begon2024back}, NMOS selectors were used to partially or totally prevent programming in "1T-1R" cells (1 transistor - 1 resistive weight).  We fabricate, characterize, and model the weight update rule in these devices, with particular emphasis on their nonlinear and state-dependent programming dynamics. Building on this characterization, we introduce a device-aware learning framework in which gradient-based updates are accumulated digitally and converted into discrete programming events only upon crossing a threshold, emulating realistic programming behavior while reducing update activity. Using this framework, we evaluate device-aware on-device learning, demonstrate subject-specific transfer learning through retraining only the final network layers, and study robustness to limited weight resolution and programming variability in weight-transfer scenarios with low-overhead device-aware re-tuning.

The remainder of this paper proceeds as follows. We first describe the motor imagery dataset, the spiking neural network architecture, and the training and evaluation pipeline, followed by the fabrication, characterization, and calibrated modeling of the ferroelectric synaptic device. We then introduce the weight-mapping procedure and the device-aware update formulation used throughout this study. Using this framework, we present results for device-aware on-device learning under thresholded programming updates, and then demonstrate subject-specific transfer learning by retraining only the final network layers. Finally, we analyze weight transfer by quantizing software-trained weights and emulating programming variability, and show that performance can be recovered with low-overhead device-aware re-tuning under realistic programming constraints.

\section{Methods}

\begin{figure}[H]
     \centering
         \includegraphics[clip,width=18cm,height=10cm,keepaspectratio, width=1\textwidth]{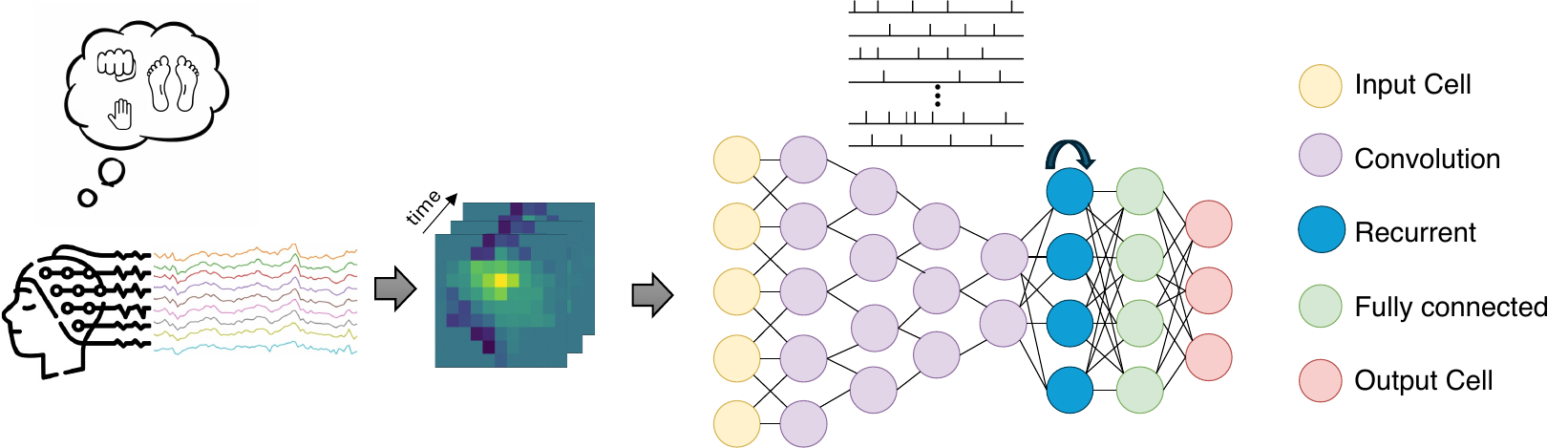}
    \caption{Spiking neural network architecture. The time series signals from all the 64 electrodes were used where the participants imagine limb movements. The electrode potential at each time step is converted to a 2D map, which is passed to the network from ref. \cite{kumar2022decoding} comprising four convolutional layers followed by a recurrent and two fully connected layers. 
}
    \label{fig:fig1}
\end{figure}

\subsection{Motor imagery dataset}

In this study, we used the Physionet \cite{goldberger2000physiobank} motor imagery dataset \cite{schalk2004bci2000}. This database comprises recording from 109 participants through a 64 channel EEG headset using BCI2000 system with a sampling rate of 160 Hz. These recordings were obtained from participants engaged in a sequence of motor and imagery activities. The tasks related to action are not considered in this study, because the model should be trained on people not able to perform actions. Same as the study by \cite{kumar2022decoding}, data from six subjects were removed, resulting in 103 subjects used in our study. The recordings are classified in two subsets of imagined actions and left hand imagery task contained 2336 trials in total and right hand imagery task contained 2299 trials in total. The signals were band-pass filtered between 0.1-80 Hz, and a 1-second segment (0-1 s after cue) was extracted for each trial.

\subsection{Spiking neural network}

Our model follows the spiking neural network (SNN) architecture proposed in~\cite{kumar2022decoding}, which we adopt because it was specifically developed for EEG motor-imagery decoding and provides a suitable balance between spatial feature extraction, temporal modeling, and network complexity. We preserve the overall topology, including the spike-encoding stage, three convolutional spiking layers (CONV1--CONV3), a temporal convolution layer, a recurrent layer, and two fully connected layers (FC1--FC2) for classification. The 64 EEG channels are projected onto a $10\times11$ spatial grid with zero padding at unoccupied locations, yielding a 2D representation that preserves the approximate scalp topography and local electrode neighborhood relationships, consistent with prior EEG spatial embedding approaches \cite{carabez2017convolutional} and standardized electrode placement \cite{acharya2016american}. The grid accommodates the full electrode arrangement while retaining empty positions where no electrode is present, thereby preserving relative spatial structure rather than forcing the channels into an artificial dense image.

Within this representation, the use of \(3\times3\) convolutional kernels allows the network to capture local spatial correlations among neighboring electrodes without introducing an excessively large number of parameters. Stacking multiple such layers progressively enlarges the effective receptive field, so that CONV1 extracts local spatial patterns, while CONV2 and CONV3 build progressively more abstract spatial features at each time step. All three convolutional layers use $3\times3$ kernels, which provide a good compromise between capturing short-range spatial correlations and limiting parameter count; stacking such layers also progressively enlarges the effective receptive field, following common deep CNN design principles \cite{simonyan2014very}. In parallel, the number of feature maps increases with depth (64, 128, and 256 for CONV1--CONV3, respectively), forming a hierarchical or pyramidal feature extractor in which earlier layers capture simpler local structure and deeper layers encode richer spatial combinations \cite{ullah2016pyramid}. After the spatial feature extractor, the temporal convolution layer (TC1) uses a kernel spanning three time steps to capture short-range temporal dynamics, while the recurrent layer models longer-range temporal dependencies across the sequence.

At each time step, the network produces a vector of classification scores. These logits are modulated by a set of learnable temporal weights $\mathbf{w}_{ts}$ and subsequently aggregated to yield the final prediction:
\begin{equation}
\mathbf{y} = \sum_{t=1}^{T} w_{ts}(t)\, \mathbf{o}(t),
\end{equation}
where $\mathbf{o}(t)$ denotes the output of the final fully connected layer at timestep $t$. This learnable temporal weighting mechanism enables the model to assign greater importance to more informative temporal segments during the decision-making process.

The complete architecture is shown in \autoref{fig:fig1}, and the layer dimensions together with the number of trainable parameters are summarized in Table~\ref{tab:architecture}.

\begin{table}[t]
\footnotesize
\centering
\caption{Layer-wise architecture with feature map sizes, trainable parameters, 
and initialization weight bounds.}
\label{tab:architecture}
\begin{tabular}{lcccccc}
\toprule
\textbf{Layer} & \textbf{Output size} & \textbf{Kernel / Type} &
\textbf{\# Units / Filters} & \textbf{Synaptic} & 
\textbf{Neuron} & \textbf{Weight bound} \\
\midrule
CONV1 & $64\times8\times9$   & $3\times3$ spatial & 64 filters & 576      & 4{,}608 & 0.3330 \\
CONV2 & $128\times6\times7$  & $3\times3$ spatial & 128 filters & 73{,}728 & 5{,}376 & 0.0417 \\
AvgPool & $128\times3\times3$ & $2\times2$ & -- & -- & -- & -- \\
CONV3 & $256\times1\times1$  & $3\times3$ spatial & 256 filters & 294{,}912 & 256 & 0.0295 \\
TC1   & $256$                & temporal kernel = 3 & 256 units & 196{,}608 & 256 & 0.0625 \\
R1    & $256$                & recurrent & 256 units & 65{,}536 & 256 & 0.0625 \\
FC1   & $256$                & linear & 256 units & 65{,}536 & 256 & 0.0625 \\
FC2   & $2$                  & linear & 2 units & 512 & 2 & 0.0625 \\
Time weights & $T$           & -- & 1 & $T$ & -- & -- \\
\midrule
\textbf{Totals} & -- & -- & -- & \textbf{697{,}408} & 
\textbf{11{,}010} & -- \\
\bottomrule
\end{tabular}
\end{table}

All spiking layers implement a leaky integrate-and-fire (LIF) neuron with trainable decay factors:
\begin{align}
i_t &= \beta\, i_{t-1} + I_t^{\text{syn}}, \\
v_t &= \gamma\, v_{t-1}(1 - s_{t-1}) + i_t, \\
s_t &= H(v_t - v_{\mathrm{th}}),
\end{align}
where $s_t\!\in\!\{0,1\}$ denotes the spike output and $\beta,\gamma$ are learned decay parameters.

These neurons are modeled using a discrete-time LIF neuron formulation. This choice is motivated by the fact that the EEG signals are temporal data, and LIF neurons intrinsically model the temporal dynamics of the signal. In addition, EEG inputs are acquired as sampled time series and are therefore naturally processed in discrete time steps, whereas surrogate-gradient training requires explicit time unrolling of neuronal dynamics during backpropagation through time. 

A rectangular surrogate gradient was used during backpropagation to approximate the derivative of the spiking activation function.
\begin{equation}
\frac{\partial s_t}{\partial v_t} =
\begin{cases}
A, & \text{if } |v_t - v_{\mathrm{th}}| < g, \\[6pt]
0, & \text{otherwise},
\end{cases}
\label{eq:rect-surrogate}
\end{equation}
where $A$ denotes the gradient amplitude, $v_t$ is the membrane potential, $v_{\mathrm{th}}$ is the firing threshold and $g$ defines the linear window width.

\subsection{Training and evaluation pipeline}

For model evaluation, we used a five-fold cross-validation procedure to ensure subject-independent testing. The EEG data set comprising 103 participants was partitioned into five disjoint subsets according to the identifiers of the subjects: the first two subsets contained 21 participants each, while the remaining three contained 20 participants each. In each fold, data from four subsets ($\approx$82 participants) were used for network optimization, including an 80/20 split for training and validation. The remaining subset (approximately 21 participants) served as an unseen test set to evaluate generalization to new subjects. The reported performance metrics correspond to the mean and standard deviation of test accuracy across all five folds.

Training was performed with an initial learning rate of 0.0001,
using a CosineAnnealing scheduler that gradually decayed the learning rate to 0.00001 by the end of training. Each model was trained for 20 epochs in the pretraining phase, and 5 epochs in the subjects specific finetuning. The batch size used for pretraining was 64 and for subject specific finetuning was 1, to simulate the scenario of online learning. The Adam optimizer was used in the training, which combines momentum (first moment $m_t$) and adaptive learning rate(second moment $v_t$) with bias correction by equation \ref{eq:bias correction} to update weight \ref{eq:Adam}:
\begin{equation}
m_t = \beta_1 m_{t-1} + (1 - \beta_1) \frac{\partial L}{\partial w_t}
\end{equation}

\begin{equation}
v_t = \beta_2 v_{t-1} + (1 - \beta_2) (\frac{\partial L}{\partial w_t})^2
\end{equation}

\begin{equation}
\hat{m}_t = \frac{m_t}{1 - \beta_1^t}, \qquad
\hat{v}_t = \frac{v_t}{1 - \beta_2^t}
\label{eq:bias correction}
\end{equation}

\begin{equation}
w_t = w_{t-1} - \alpha \frac{\hat{m}_t}{\sqrt{\hat{v}_t} + \epsilon}
\label{eq:Adam}
\end{equation}

\subsection{Synaptic device}

A ferroelectric synaptic device with the same functional stack as that of \cite{begon2024back} was fabricated on a silicon subtrate.
The substrate is a 0.8 \textmu m thick thermal SiO\textsubscript{2} on Si chip. Using Atomic Layer Deposition (ALD), 20 nm of TiN was deposited at 300\textdegree C, then 45 cycles of WO\textsubscript{x} at 360\textdegree C. Afterwards, a [HfO\textsubscript{2}/ZrO\textsubscript{2}] nanolaminate is deposited at 300\textdegree C consisting of five supercycles, each comprising five cycles with tetrakis (ethylmethylamino) hafnium (IV) and O\textsubscript{2}, and ten cycles with bis (methylcyclopentadienyl) (methyl) (methoxy) zirconium (IV) and O\textsubscript{2}.
Ten nanometers of TiN were deposited in situ. The ALD deposition was followed by a 50 nm of tungsten sputtering. Crystallization was performed with a millisecond flash lamp anneal: the sample was preheated at 450\textdegree C for 120 s, followed by a 20 ms pulse of 90 J/cm\textsuperscript{2}. The top electrode was defined by UV lithography and reactive ion etching (RIE). 
A 100 nm thick SiO\textsubscript{2} passivation layer was then sputtered. A cross-section is represented in \autoref{fig:figx-device}(a).
-

\begin{figure}[H]
     \centering
         \includegraphics[clip,width=18cm,height=10cm,keepaspectratio, width=1\textwidth]{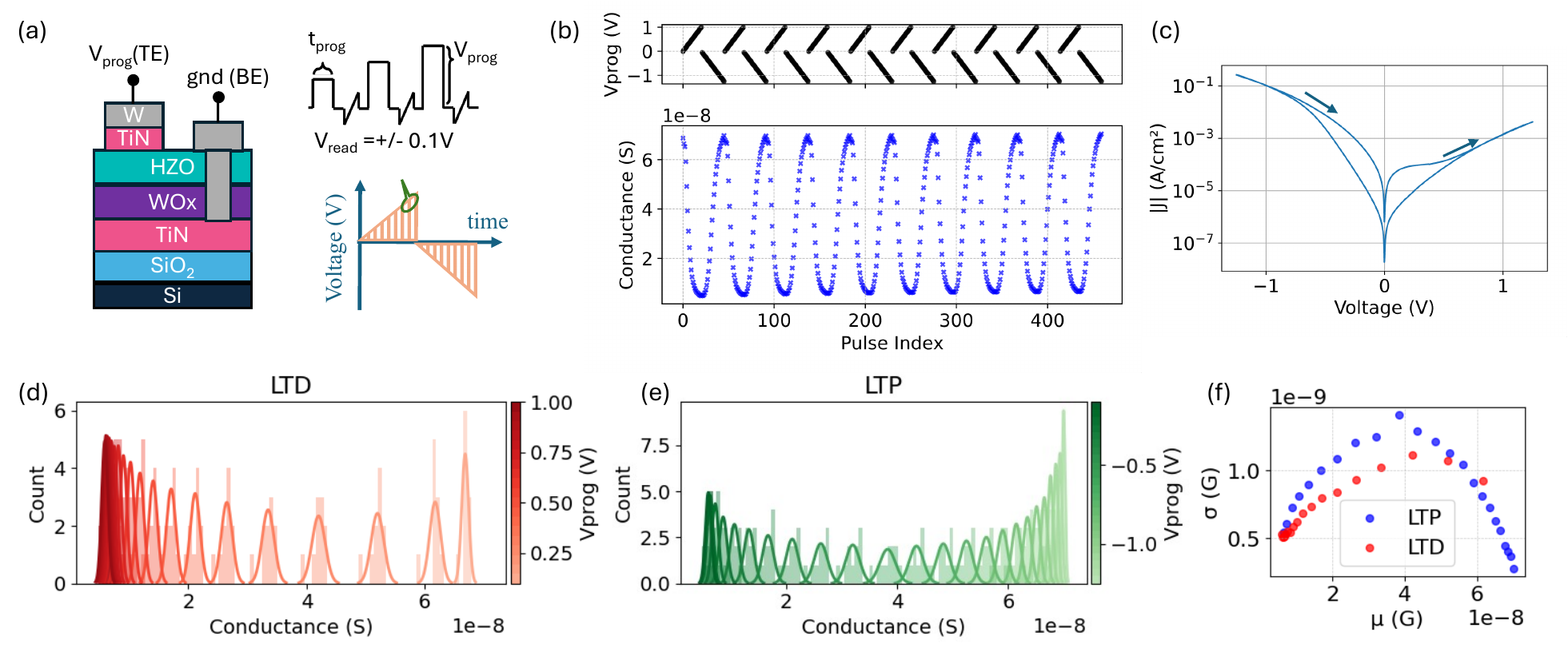}
    \caption{Ferroelectric synaptic device programming. 
    {\footnotesize \textbf{a} Schematic of the ferroelectric synaptic device stack and the programming scheme, consisting of write pulses with fixed pulse width (50~\textmu s) and increasing amplitudes (\(V_{\mathrm{prog}}\)). Positive-polarity pulses induce long-term depression (LTD), followed by negative-polarity pulses that induce long-term potentiation (LTP). The device conductance is read after each programming pulse. \textbf{b} Characterization results showing the applied programming-pulse amplitudes (top) and the corresponding evolution of device conductance over time (bottom) for ten programming cycles. \textbf{c} Current density–voltage (J-V) characteristics of the synaptic device, showing the forward and reverse sweep directions. Programmed conductance values grouped by pulse amplitude and fitted with Gaussian distributions for LTD (\textbf{d}) and LTP (\textbf{e}) pulse sequences. \textbf{f} Standard deviation of the fitted conductance distributions plotted as a function of the programmed conductance level. 
}}
    \label{fig:figx-device}
\end{figure}

To characterize multiple conductance states,  a 8300~\textmu m\textsuperscript{2} device was repeatedly cycled between high and low-resistance states (HRS/LRS) applying a sequence of programming pulses, as shown in \autoref{fig:figx-device}(b). The programming pulses had a constant width of 500~\textmu s and their amplitude gradually increased from 0 V to +1 V for potentiation and from 0 V to -1.25 V for depression, in +/-50 mV steps. For read operations, bipolar triangular pulses of +/-100 mV were applied, and the resulting current was measured using a source-measure unit (SMU). The measured resistance values ranged from approximately 10 MΩ to 100 MΩ. The physical mechanisms leading to a change in conductance upon programming pulses are detailed in ref.\cite{begon2024back}: at each pulse, the direction of the polarization for a fraction of the ferroelectric domains flips. The screening of the polarization results in a local redistribution of charges, modifying the conduction through the insulator. In addition, a current density–voltage 
(J-V) sweep was performed to obtain the characteristic hysteresis shown in \autoref{fig:figx-device}(c).

The programmable conductance states were grouped according to the magnitude of the programming pulses, as shown in \autoref{fig:figx-device}(d-e) for long-term depression (LTD) and long-term potentiation (LTP).

\subsection{Weight mapping}


All synaptic weights of network $w_{\mathrm{SNN}}$ can be mapped to a normalized differential representation compatible with the memristive crossbar. For a given layer with fan-in of $\mathrm{fan}_{\mathrm{in}}$, we define the initialization bound as
\begin{equation}
    \mathrm{bound} = \frac{1}{\sqrt{\mathrm{fan}_{\mathrm{in}}}},
\end{equation}
such that the trained weights typically satisfy $w_{\mathrm{SNN}} \in [-\mathrm{bound}, \mathrm{bound}]$, consistent with the fan-in-based Kaiming He weight initialization scheme commonly used in deep neural networks~\cite{he2015delving}. During all the experiments including the noise addition and on-device learning, the weights were clamped to this range. These weights were then rescaled into a differential weight
\begin{equation}
    w_{\mathrm{diff}} = \frac{1}{2}\,\frac{w_{\mathrm{SNN}}}{\mathrm{bound}},
\end{equation}
which lies in the interval $w_{\mathrm{diff}} \in [-0.5, 0.5]$.
We define $w_{\mathrm{diff}}$ as the difference between two normalized memristive
conductances, $w_{\mathrm{diff}} = w^{+} - w^{-}$, and fix the reference device to
\begin{equation}
    w^{-} = 0.5,
\end{equation}
so that the programmed value of the active device becomes
\begin{equation}
    w^{+} = w_{\mathrm{diff}} + 0.5
          = \frac{1}{2}\,\frac{w_{\mathrm{SNN}}}{\mathrm{bound}} + 0.5,
\end{equation}
with $w^{+} \in [0,1]$.

Bipolar weights are represented using a differential-pair scheme in which the complementary device is fixed at mid-level conductance, while the active device alone encodes the signed weight. Based on retention measurements reported in \cite{begon2024back}, the programmed mid-level conductance state was found to be stable over time. Consequently, the hardware update model is applied only to the active device.  

\subsection{Device modeling and weight-update formulation}

\begin{figure}[H]
     \centering
         \includegraphics[clip,width=18cm,keepaspectratio, width=0.7\textwidth]{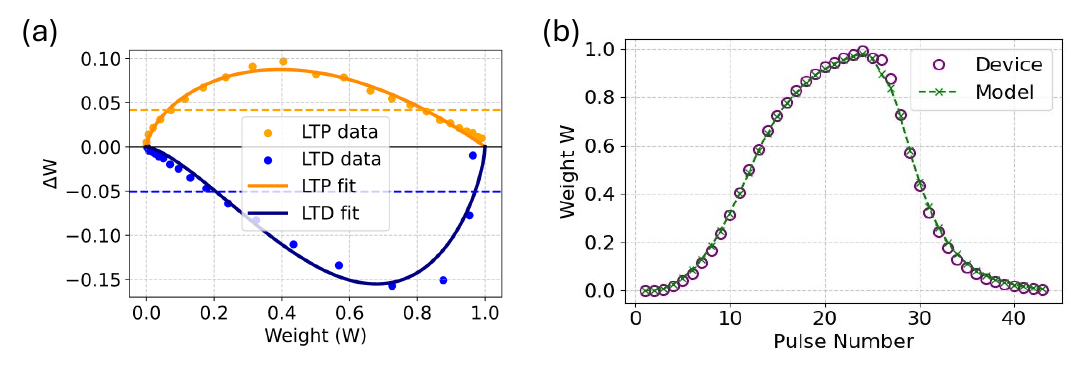}
    \caption{Model
    {\footnotesize \textbf{a} The change in normalized weight ($\Delta W$) is plotted with respect to the initial weight for long-term potentiation (LTP) and depression (LTD) including the device characterization data and the fitted model. The dotted lines corresponds to the mean value of ($\Delta W$) obtained through the device characterization data for LTP and LTD. \textbf{b} The evolution of weight with programming pulse number is plotted for the device data and model.}
    }
    \label{fig:model}
\end{figure}

The weight updates on-device learning is modeled using a phenomenological conductance update law. Based on prior measurements of ferroelectric synaptic devices, the incremental update of the normalized synaptic weight $W \in [0,1]$ after a programming pulse is expressed as a scaled Beta-shaped kernel:

\begin{equation}
    \Delta W_{\mathrm{LTP}}(W) = A_{+}\, W^{\alpha_{+}-1}(1-W)^{\beta_{+}-1}, \qquad
    \Delta W_{\mathrm{LTD}}(W) = -A_{-}\, W^{\alpha_{-}-1}(1-W)^{\beta_{-}-1}.
    \label{eq:beta-kernel}
\end{equation}

This form captures the experimentally observed nonlinear and asymmetric switching behavior of ferroelectric synaptic devices, as shown in \autoref{fig:model}. The three parameters $(A, \alpha, \beta)$ control the update magnitude and shape. They are determined through least-squares fitting to device characterization data: 

\begin{align*}
    A_{+} &= 0.1761, & \alpha_{+} &= 1.81, & \beta_{+} &= 2.12,\\
    A_{-} &= 0.3300, & \alpha_{-} &= 2.47, & \beta_{-} &= 1.79.
\end{align*}

\subsection{On-device learning}

\begin{figure}[H]
     \centering
         \includegraphics[clip,width=18cm,keepaspectratio, width=0.7\textwidth]{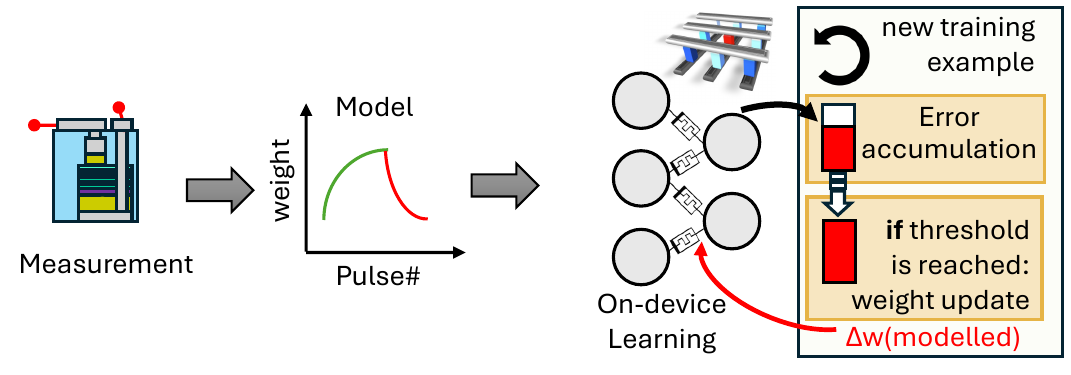}
    \caption{Simulation framework: The measurements from device characterization was used to fit the model. Thereafter the model is used to compute the weight updates of the network. The weight updates computed from the gradient descent is accumulated until a predefined threshold is surpassed, after which the weight update is computed through the fitted model and applied to the respective weights. 
    }
    \label{fig:framework}
\end{figure}

During online learning, each synaptic weight maintains an accumulated update $\delta w_{\mathrm{acc}}$ using backpropagated gradients (\autoref{fig:framework}). The hardware update is not performed as long as $|\delta w_{\mathrm{acc}}| < \varepsilon$, where $\varepsilon$ is a threshold that represents the minimum change that can be reliably programmed in the device. This accumulation-based update mechanism is conceptually similar to the mixed-precision learning strategy previously proposed for non-spiking networks with memristive~\cite{nandakumar2020mixed} and ferroelectric synapses~\cite{garg2025energy}.

Once $|\delta w_{\mathrm{acc}}| \ge \varepsilon$, a single programming event is triggered. The sign of $\delta w_{\mathrm{acc}}$ determines whether a potentiation (LTP) or depression (LTD) pulse is applied to the active device. The resulting conductance change is computed from the Beta-kernel based device model. After the update, $\delta w_{\mathrm{acc}}$ is reset to zero. The updated device conductance $W$ is then mapped back to the neural weight domain through the scaling described earlier. During the on-device learning phase, each synaptic weight maintains an accumulated update $\delta w_{\mathrm{acc}}$ based on the backpropagated gradient. After each batch (64 trials), this accumulator is compared against asymmetric LTP/LTD thresholds, and the device update is chosen according to
\begin{equation}
\Delta W =
\begin{cases}
\Delta W_{\mathrm{LTP}}(W), & \text{if } \delta w_{\mathrm{acc}} \ge \varepsilon, \\[6pt]
0,                          & \text{if } -\varepsilon\,\varepsilon_{\mathrm{asym}}
                               < \delta w_{\mathrm{acc}} < \varepsilon, \\[6pt]
\Delta W_{\mathrm{LTD}}(W), & \text{if } \delta w_{\mathrm{acc}} \le
                               -\varepsilon\,\varepsilon_{\mathrm{asym}}.
\end{cases}
\label{eq:acc-threshold}
\end{equation}

Here $\varepsilon$ is the base threshold, defined as a percentage of the fan-in-dependent weight range of each layer, and
$\varepsilon_{\mathrm{asym}}$ scales the LTD threshold to reflect the larger step sizes observed for depression compared to potentiation in the device measurements. Once an LTP or LTD programming event is triggered, the corresponding conductance change is computed from the beta-shaped kernel device model in Eq.~\eqref{eq:beta-kernel}. After the update, $\delta w_{\mathrm{acc}}$ is reset to zero, and the updated device conductance $W$ is mapped back to the neural weight domain via the scaling described earlier.

With the chosen accumulation thresholds, each weight can be updated at most once per batch. In practice, we observed that even when the threshold was set to only $5\%$ of the fan-in-based range, the individual backpropagation updates were still too small to cause multiple LTP/LTD events for the same weight within a single batch. Therefore, accumulation was performed across epochs, so that many small gradient updates could still build up and eventually trigger discrete, hardware-like weight changes.

\subsection{Subject specific transfer learning}

Subjects excluded from each pretraining fold were the target subjects for fine tuning of the network. The pre-trained model, trained with the fully on-device learning method was used for subject-specific fine-tuning experiments. For each participant, the trials were divided into four equal subsets. In each fold, three subsets were used for fine-tuning, while the remaining subset was used solely for evaluation. Due to the fewer number of trials for each participant, and to emulate the online learning scenario, a batch size of one was used for this fine-tuning. Moreover, this fine-tuning was performed with the fitted memristor model. After cycling through all four folds, classification accuracy was calculated for each fold, and the overall accuracy for the participant was obtained by averaging the accuracies across the four folds.

\section{Results and Discussion}

\subsection{Baseline}

\begin{figure}[H]
     \centering
         \includegraphics[clip,width=18cm,keepaspectratio, width=0.3\textwidth]{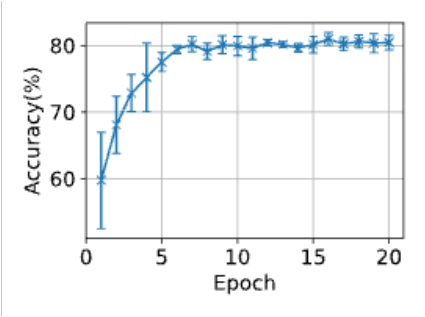}
    \caption{Software baseline: Network's classification performance for 2-class problems: Left/Right hand classification on the validation set plotted against the number of epochs. The error bars depicts the standard deviation across five folds.
    }
    \label{fig:baseline}
\end{figure}

We first reproduced the floating-point training baseline of Kumar et al.~\cite{kumar2022decoding} using the same EEG motor-imagery dataset and network architecture, implemented within our spiking neural network framework. The validation accuracy from all five folds is plotted across training epochs in \autoref{fig:baseline}. The reproduced model achieved a test accuracy of  80.39~$\pm$~2.98\% for the two-class motor imagery task (left vs.~right hand), closely matching the performance of 80.65 $\pm$ 3.83\% reported earlier in ~\cite{kumar2022decoding} for spiking neural network. For non-spiking neural networks, \cite{dose2018end} reported an accuracy of 82.43\% using 4-s EEG segments, while \cite{wang2020accurate} achieved 80.38\% using the first 2 s of each trial. In contrast, as described in the Methods section, our approach considers only the first 1 s of the EEG segment for classification. This design choice is motivated by the need to substantially reduce intent-to-command latency, which is critical for real-time BCI operation.

\subsection{On-device learning}

\begin{figure}[H]
     \centering
         \includegraphics[clip,width=18cm,keepaspectratio, width=1\textwidth]{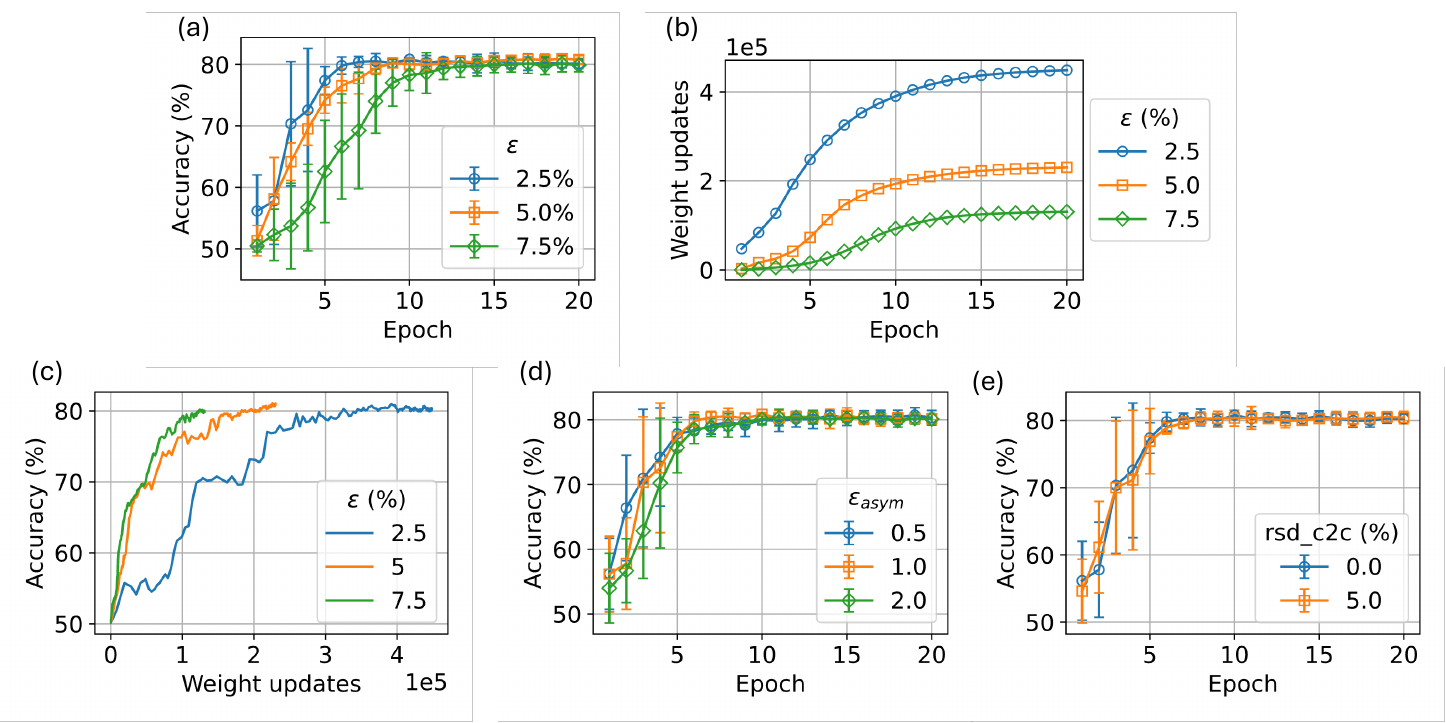}
    \caption{On-device learning 
    {\footnotesize \textbf{a} Validation accuracy of the network as a function of training epochs for different update thresholds (\(\epsilon\)) of 2.5\%, 5\%, and 7.5\%. 
    \textbf{b} Total number of weight updates accumulated across all synapses during training for the corresponding thresholds.
    \textbf{c} Validation accuracy plotted with respect to the total number of weight updates for different update thresholds. 
    \textbf{d} Validation accuracy for asymmetric update thresholds (\(\epsilon_{\mathrm{asym}}\)) of 0.5, 1, and 2, where \(\epsilon_{\mathrm{asym}} = \epsilon_{+}/\epsilon_{-}\) denotes the ratio between potentiation and depression thresholds.
    \textbf{e} Validation accuracy during pre-training when cycle-to-cycle variability is included in the device-update model by sampling the update amplitude ($A_{+}$ and $A_{-}$) with a relative standard deviation of 5\%, compared with the nominal case without variability.
    }
    }
    \label{fig:online_learning}
\end{figure}

We trained the network from an untrained initialization using the threshold-based update triggering scheme, where the weight updates were computed using the fitted device model. The results for three different values of the update threshold (\(\epsilon\)) are shown in \autoref{fig:online_learning}(a). In this experiment, the asymmetry factor was fixed to $\varepsilon_{asym} = 1$ to isolate the effect of the relative threshold $\varepsilon$. The threshold values of 2.5\%, 5\%, and 7.5\% were chosen to lie in the same range as the experimentally observed programming granularity of the ferroelectric device. From the fitted device-update characteristics (see \autoref{fig:model}a), the programming increments span from below 1\% up to approximately 10\% for LTP and 15\% for LTD depending on the conductance state, while the mean update magnitudes are around 4\% for LTP and 5\% for LTD. The selected thresholds therefore probe a practically relevant regime in which the triggering condition is comparable to the typical device programming step size. A higher threshold corresponds to a coarser effective weight granularity, leading to less frequent device updates and slower convergence during training. Increasing the update threshold reduces the cumulative number of synaptic programming events required to reach a given classification accuracy, highlighting a trade-off between learning accuracy, convergence speed, and programming activity. As shown in \autoref{fig:online_learning}(b) and (c), higher thresholds reach similar accuracy with fewer weight updates but require more learning iterations, resulting in lower programming energy and improved device lifetime. The test accuracies after training are $79.52\% \pm 3.15\%$, $79.13\% \pm 2.1\%$ and $78.91\% \pm 2\%$ for the threshold values of 2.5\%, 5\% and 7.5\%, respectively.

\autoref{fig:online_learning}(d) further shows the impact of asymmetric update thresholds (\(\epsilon_{\mathrm{asym}}\)) on the validation accuracy. For this sweep, we set \(\varepsilon = 2.5\%\). The test accuracies are \(79.92\% \pm 2.5\%\), \(79.52\% \pm 3.15\%\), and \(78.91\% \pm 2.76\%\) for asymmetry factors of 0.5, 1, and 2, respectively. \(\epsilon_{\mathrm{asym}}\) of 0.5 shows the fastest convergence, and this is due to the fact that the LTD had steeper updates (A- = 0.033, A+ = 0.171), and thus this asymmetry in switching is compensated by the asymmetric thresholds. Nevertheless, the network exhibits convergence for all three degrees of asymmetry, showcasing the robustness of learning against asymmetric update behaviour. These results show that using a memristor model with a discrete number of levels and non-linear weight update, the accuracy of the network is comparable to the digital baseline. Furthermore, to account for cycle-to-cycle variability representative of the physical device, we sampled the update amplitude in the Beta-shaped kernel from a normal distribution with a relative standard dispersion of 5\%. From \autoref{fig:figx-device}(f), the programmed weight variation in the characterized device reaches a maximum relative standard dispersion of 3.75\%. In this way, each programming event during training experiences a slightly different effective update magnitude, thereby emulating cycle-to-cycle fluctuations in the ferroelectric synapse. As shown in \autoref{fig:online_learning}(e), the convergence trajectory remains very similar to the nominal case, and the final test accuracy after 20 epochs of training reaches $79.60\% \pm 2.92\%$ with cycle-to-cycle variations included. This indicates the robustness of the proposed training framework to stochastic programming variations observed in the device.

\subsection{Subject-specific transfer learning}

\begin{figure}[H]
     \centering
         \includegraphics[clip,width=18cm,keepaspectratio, width=0.8\textwidth]{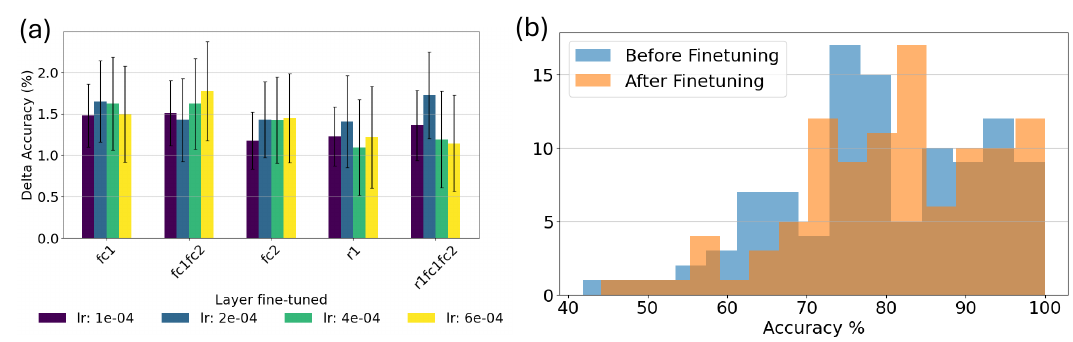}
    \caption{Subject specific transfer learning 
    {\footnotesize \textbf{a} Test accuracy across all participants when re-training different subsets of network layers using four learning rates. Error bars denote the standard error across participants. \textbf{b} Distribution of participant-wise test accuracy before and after fine-tuning. Only the fc1 and fc2 layers were fine-tuned with learning rate of 6e-04. 
}}
    \label{fig:sstl}
\end{figure}

EEG signals have a strong subject-specificity. In this section, we explore the feasibility of a subject-specific transfer learning (SSTL) strategy using ferroelectric synapses. In SSTL, a network pre-trained across four folds of participants and then subsequently fine-tuned for each individual of the fifth fold. The pre-trained model corresponds to a fully on-device trained model with \(\epsilon\) of 2.5\% and \(\epsilon_{\mathrm{asym}}\) of 1, for a simpler implementation. These parameters and the memristive model was used for SSTL.  Rather than retraining the full network, adaptation is restricted to selected subsets of layers in order to avoid underfitting due to a limited training data of only one subject. \autoref{fig:sstl}(a) compares the effect of retraining different layer groups using multiple learning rates. Transfer learning studies have shown that fine-tuning only higher-level layers often provides better generalization than retraining the full network, particularly when the target dataset is limited. Updating a large fraction of the network parameters can lead to overfitting or degradation of previously learned representations \cite{yosinski2014transferable, li2017learning}. Based on this analysis, fine-tuning was restricted to the final fully connected layers (fc1 and fc2) using a learning rate of $6\times10^{-4}$. The final test accuracy cumulated for all the participants, was 81.33\%, corresponding to an improvement of 1.77\% over the pre-trained network. \autoref{fig:sstl} (b) shows the participant-wise test accuracy before and after subject-specific fine-tuning under this configuration. A systematic improvement in accuracy is observed for the majority of participants, despite a finite number of conductance levels for the ferroelectric weights and a non-linear update rule. This demonstrates that using ferroelectric synapses, personalization at the classifier level is sufficient to capture individual-specific EEG characteristics without modifying the full network.

\subsection{Weight transfer and re-tuning}


\begin{figure}[H]
     \centering
         \includegraphics[clip,width=18cm,keepaspectratio, width=0.7\textwidth]{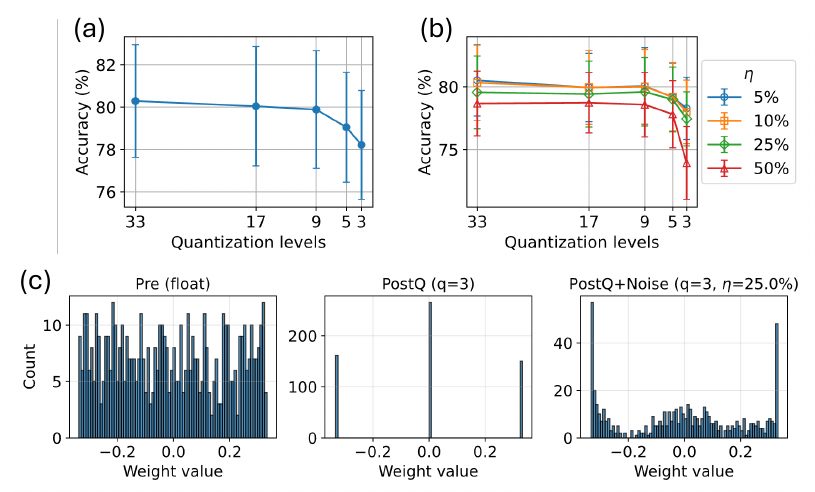}
    \caption{Quantization of weights 
    {\footnotesize \textbf{a} Test accuracy versus quantization levels. The error bars depicts the standard deviation across five folds. \textbf{b} Test accuracy for different quantization levels and additive noise ($\eta$). \textbf{c} The histogram of weights of Conv1 layer after training, with quantization to 3 levels, and addition of noise of $\eta$=25\%. 
}}
    \label{fig:quant}
\end{figure}

In this section, instead we investigate to which extent a transfer learning strategy can be implemented with ferroelectric synapses. The starting point is the baseline network (see \autoref{fig:baseline}) trained with floating-point precision. First, the effect of quantization on the network's performance is measured. Synaptic weights were quantized to a specified number of discrete levels using uniform binning. For the two-class motor imagery classification task, as shown in \autoref{fig:quant} (a) the accuracy remains as high as 78\% even when the weights are constrained to only three quantization levels ($-\max$, $0$, $+\max$). This robustness to coarse weight quantization is a well-known strength \cite{li2020robustness} of spiking neural networks, in which information is primarily encoded and propagated through spike timing rather than precise synaptic weight values \cite{rueckauer2017conversion}. Consequently, the burden of precision is shifted from the weight domain to the temporal domain (e.g., spike times), enabling tolerance to low-resolution synaptic weights.

To emulate device-level programming variability, Gaussian noise was added to the quantized weights. The noise standard deviation was defined as a fraction $\eta$ of the mean value of the non-zero quantized levels. The relationship between the network’s test accuracy and quantization levels was evaluated for $\eta$ ranging from 5\% to 50\%, the corresponding results are presented in \autoref{fig:quant}(b). For the characterized device, the maximum $\eta$ of programmed weight from \autoref{fig:figx-device}(f) is 3.75\%. These measurements correspond to a single device and we can anticipate a higher programming error when accounting for device-to-device variability. Nevertheless, from the results, we can infer that only a drop in accuracy of 2\% is observed for an extreme noise level of 25\%. Prior work \cite{shen2024conventional, stromatias2015robustness} has shown that SNNs can retain accuracy under substantial synaptic-weight noise and reduced precision, and in some regimes can be more robust to weight perturbations than comparable non-spiking neural networks \cite{li2020robustness}, supporting the use of low-resolution (e.g., low-bit or binary) weights for neuromorphic deployment. While these results highlight the robustness of the adopted SNN to coarse weight quantization and programming noise, a direct comparison with a non-spiking network under the same architectural and ferroelectric hardware-mapping constraints remains an important direction for future work.

\begin{figure}[H]
     \centering
         \includegraphics[clip,width=18cm,keepaspectratio, width=1\textwidth]{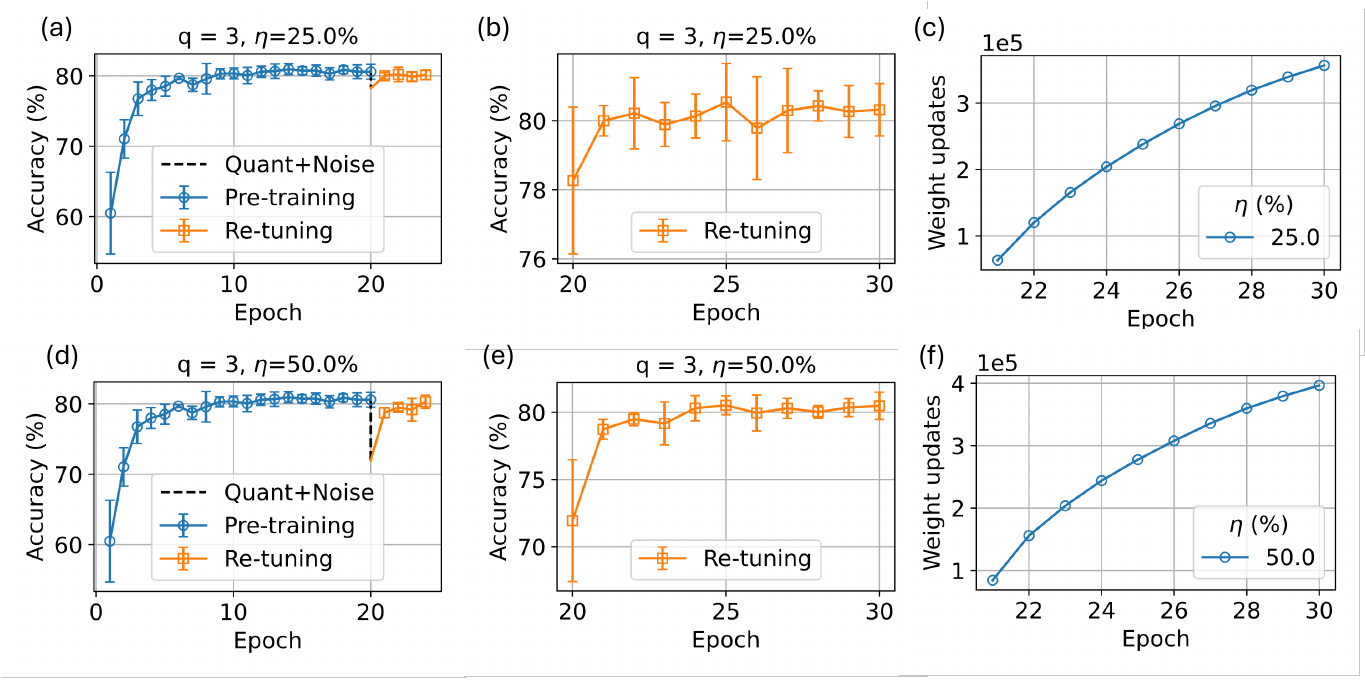}
    \caption{Re-tuning of quantized weights.
    {\footnotesize Validation accuracy across epochs is shown for the pre-training phase, followed by quantization to three levels and additive noise with standard deviation $\eta=25\%$ (\textbf{a}) and $\eta=50\%$ (\textbf{d}), and subsequent re-tuning using the memristive device model. Panels (\textbf{b}) and (\textbf{e}) show the re-tuning phase extended to 10 epochs, while panels (\textbf{c}) and (\textbf{f}) report the cumulative number of weight updates during re-tuning for the corresponding noise levels.
}}
    \label{fig:fine-tune}
\end{figure}

Training a network entirely on hardware can significantly impact device endurance due to frequent programming operations. Therefore, we focus on a deployment strategy based on weight transfer followed by limited on-device re-tuning. In this approach, the network weights are first quantized and perturbed to account for programming variability, and the resulting network is subsequently fine-tuned for a small number of epochs using the memristive weight-update model. This setup emulates a practical open-loop deployment scenario, in which weights trained in software are transferred to a memristive crossbar and adapted on-device under realistic programming constraints. For three quantization levels and an additive noise level of $25\%$ and $50\%$, the validation accuracy recovers to approximately $80\%$ within a single training epoch, as shown in \autoref{fig:fine-tune}. The weight-update threshold in the accumulator was set to $\epsilon = 2.5\%$. After re-tuning, we obtain a test accuracy of $79.08\% \pm 2.16\%$ for $\eta = 25\%$ and $78.84\% \pm 2.39\%$ for $\eta = 50\%$, which in both cases represents only a slight decrease of around 1.5 percentage points relative to the test accuracy after pre-training in floating-point precision. An extended re-tuning the network for 10 epochs yields only limited additional improvement compared with the results obtained after 4 epochs. For $\eta = 25\%$, the test accuracy increases only slightly from $79.08\% \pm 2.16\%$ to $80.46\% \pm 2.63\%$, and for $\eta = 50\%$, from $78.84\% \pm 2.39\%$ to $79.50\% \pm 2.91\%$. Rather than the rapid recovery observed during the first one to two re-tuning epochs, the validation accuracy fluctuates around a similar level after the fourth epoch. Meanwhile, the cumulative number of weight updates still increases, but progressively more slowly, indicating diminishing returns in accuracy relative to additional programming activity. This supports the choice of limiting the re-tuning phase to only a few epochs.


\subsection{Hardware implications}

Beyond the algorithmic results, the adopted bipolar weight mapping also has important hardware implications. In the present scheme, signed weights are represented using a differential-pair configuration in which one device is fixed at a mid-level conductance and the other encodes the effective weight. This differs from conventional differential encoding, commonly used in phase-change memory (PCM) \cite{pistolesi2024differential} and other device technologies, where one device stores the positive and the other the negative contribution of a weight, often motivated by strongly asymmetric potentiation and depression characteristics. In contrast, ferroelectric devices exhibit granular and bidirectional conductance modulation, enabling both LTP and LTD on a single device. This alleviates conductance saturation and reduces the refresh operations often required in two-device differential schemes \cite{nandakumar2020mixed}. In addition, this approach requires only one column of devices as a reference column rather than duplicating the full array. This reduces hardware overhead compared with fully differential signed-weight mappings, in which two programmable devices or columns are required per weight. In the present scheme, the remaining cost is mainly associated with the reference column and the corresponding subtraction or readout circuitry, rather than a second fully programmable branch. As a result, both area and programming energy can be reduced relative to conventional two-device differential implementations while still supporting bipolar weight representation. Offset-based bipolar synapses have also been demonstrated in hardware using digital offset correction \cite{meng2021digital}, in online learning with magnetic tunnel junctions \cite{mondal2019situ}, and through reference columns to mitigate sneak-path effects \cite{fouda2019mask}.

At the array level, the conductance range considered here also appears compatible with the layer dimensions studied in this work. Guidelines for estimating the maximum array size were proposed by Gokmen and Vlasov \cite{gokmen_acceleration_2016}: for a typical line resistance of $r = 0.36~\Omega.\mu\mathrm{m}^{-1}$, the resistance of a line containing 512 synaptic weights should remain smaller than 10\% of the resistance of a single synaptic weight. In a worst-case scenario, the synaptic weights have an ON-state read resistance of $200~\mathrm{M}\Omega$ and are separated by a pitch of $100~\mu$m, corresponding to a line resistance of $18.4~\mathrm{M}\Omega$. For smaller devices, the pitch decreases and the device resistance increases, which would allow a larger number of devices per array. In terms of parasitics, for this worst-case example of 512 devices separated by a $100~\mu$m pitch, the RC delay of the metal line can be estimated as $\tau = 189$~ns using a typical parasitic capacitance of $c = 0.2~\mathrm{fF}.\mu\mathrm{m}^{-1}$. Gokmen and Vlasov further suggest using a pulse width at least ten times longer than this delay, corresponding here to more than $2~\mu$s.

In contrast to the synaptic model, which is tied to experimentally characterized device behavior, the neuron in this study is treated at the level of functional modeling rather than circuit implementation. The discrete-time LIF dynamics are therefore used as an abstract computational model and do not imply a specific circuit-level realization. Rather, the LIF equations define the temporal behavior of the network, whereas the hardware contribution of this study focuses on the ferroelectric synaptic devices used to store and update the network weights under realistic programming constraints. From a deployment perspective, this formulation is compatible with mixed-signal implementations \cite{garg2024versatile} in which ferroelectric devices implement programmable synaptic memory, while neuron-state updates such as integration, leakage, reset, and recurrence are handled either digitally or via custom analog neuron circuits.

In this work, we adopt surrogate-gradient descent with threshold-triggered programming as a device-aware training interface, rather than as a claim of fully hardware-native synaptic learning. This choice is motivated by the target application: EEG motor-imagery decoding is a supervised classification problem, for which gradient-based optimization provides a direct and well-established way to optimize task-level accuracy in deep spiking networks. Our goal here is therefore not to propose a new local ferroelectric learning rule, but to evaluate how supervised SNN training and post-deployment personalization can be constrained by experimentally measured ferroelectric programming dynamics.

The thresholded accumulation scheme should therefore be interpreted as an interface between supervised gradient-based optimization and the physical programming constraints of the ferroelectric synapse, rather than as a replacement of the device dynamics by a purely digital learning rule. The threshold determines only when a programming event is invoked; the actual conductance update is still obtained from the fitted beta-kernel device model and thus reflects the experimentally measured state- and polarity-dependent response of the device.

\subsection{Energy and feasibility}

To provide a first-order indication of the potential energy efficiency of the proposed approach, we estimated the inference and programming energy using measured device characteristics together with literature-reported values for ultra-low-power neuron circuits.

During inference, the mean effective MAC count per sample is 1,466,355 when averaged over 45 trials for subject~1, due to sparse activity. Using the measured maximum conductance of the characterized device, $G_{\max}=6.5\times10^{-8}~\mathrm{S}$, together with a read voltage of $100~\mathrm{mV}$ and an integration time of $500~\mu\mathrm{s}$, the worst case synaptic energy per MAC can be estimated as

\begin{equation}
E_{\mathrm{MAC}} \approx V_{\mathrm{read}}^2 G t_{\mathrm{int}}
= (0.1)^2 \times 6.5\times10^{-8} \times 5\times10^{-4}
= 3.25\times10^{-13}~\mathrm{J}
= 0.325~\mathrm{pJ}.
\end{equation}
This yields an estimated synaptic inference energy of
\begin{equation}
E_{\mathrm{syn}} = 1{,}466{,}355 \times 0.325~\mathrm{pJ}
\approx 4.77\times10^{-7}~\mathrm{J}
= 0.477~\mu\mathrm{J}.
\end{equation}

We further estimated the neuron energy using the ultra-low-power neuron reported in \cite{sourikopoulos20174}, assuming $4~\mathrm{fJ}$ per spike dynamic energy and $100~\mathrm{pW}$ static power:
\begin{equation}
E_{\mathrm{neuron}}
= P_{\mathrm{static}} \, T_{\mathrm{sample}} \, N_{\mathrm{neurons}}
+ E_{\mathrm{spike}} \, N_{\mathrm{spikes}}.
\end{equation}
Using a mean spike count of 569,438 over a $1~\mathrm{s}$ sample and a total of 11,010 neurons, the static neuron energy is estimated to be $1.101~\mu\mathrm{J}$, and the dynamic neuron energy is estimated to be $0.0023~\mu\mathrm{J}$, resulting in a total neuron energy of $1.103~\mu\mathrm{J}$. The total estimated inference energy per sample is therefore
\begin{equation}
E_{\mathrm{inf}} = E_{\mathrm{syn}} + E_{\mathrm{neuron}} \approx 1.580~\mu\mathrm{J},
\end{equation}
and is dominated by the static neuron contribution. As a reference point, Kumar et al. reported an inference energy of $3.60~\mathrm{mJ}$ on Intel's Loihi for a reduced version of the same broad SNN architecture, without adaptive thresholds or trainable decay parameters due to hardware constraints \cite{kumar2022neurophysiologically}. While this value provides a useful system-level hardware reference, it is not directly comparable to the present estimate, which is derived from literature-reported device- and circuit-level figures of merit rather than from a fabricated end-to-end implementation, and does not include peripheral, communication, memory-access, or control overheads. While this comparison is only indicative, it is consistent with the broader promise of analog in-memory computing for energy-efficient neuromorphic processing.

Finally, for device programming, we estimate from \autoref{fig:online_learning}(c) that the total number of weight-update events is on the order of $10^5$ for an update threshold of $7.5\%$. From the $J$--$V$ sweeps in \autoref{fig:figx-device}(c), the maximum absolute current density is $0.2660~\mathrm{A/cm^2}$ at a programming voltage of $-1.25~\mathrm{V}$. Using the device area of $8300~\mu\mathrm{m}^2$ and a programming pulse width of $500~\mu\mathrm{s}$, the programming energy can be estimated as
\begin{equation}
    E_{\mathrm{pulse}} = J A V t_p
\end{equation}

which yields an energy per programming pulse of approximately $13.8~\mathrm{nJ}$, corresponding to a total programming energy of approximately $1.38\times10^{-3}~\mathrm{J}$ ($1.38~\mathrm{mJ}$) for $10^5$ update events. However, this estimate corresponds to the present measurement conditions, namely a device area of $8300~\mu\mathrm{m}^2$ and a programming pulse width of $500~\mu\mathrm{s}$, and is therefore expected to be substantially higher than in scaled devices operated with shorter pulses. Recent work has shown that, by scaling ferroelectric hafnia devices below $100~\mu\mathrm{m}^2$, reliable programming with $20$~ns pulses can be achieved with a maximum energy of about $3$~pJ per update \cite{baigol2026analog}. These results indicate that the programming energy of the present device could be considerably reduced by lateral scaling and shorter pulse operation.

These estimates account only for the electrical energy dissipated in the device during inference and programming and do not include the energy required for gradient computation, peripheral circuitry, or the generation of programming pulses, for example, through a DAC. Moreover, the adaptation considered here is not intended as a lifelong continuous-learning scenario, but rather as a one-time or infrequent post-deployment personalization step, so the associated programming cost should be interpreted as a bounded adaptation overhead. Overall, we view the present study as a validation at the algorithmic and device-aware modeling level, while a full end-to-end hardware implementation and direct system-level measurement of inference and adaptation energy, including peripheral and control overheads, remain important directions for future work.

\subsection{Outlook}

A complete hardware-aware deployment study should also account for crossbar-level non-idealities beyond the device-update characteristics considered here, including IR drop, sneak-path currents, and half-select disturb. These effects were not explicitly modeled in the present work, but in practical implementations they can be mitigated through bounded array dimensions, selector-based cells such as 1T1R or 1S1R, and appropriate write/verify schemes \cite{jain2025heterogeneous}. In addition, such effects can be incorporated at the algorithm level through hardware-aware training, calibration, and post-programming fine-tuning using realistic crossbar models \cite{rasch2023hardware}. Extending the present framework in this direction is an important next step toward fully hardware-issues-aware neuromorphic processing.

An important next step is to move from supervised calibration toward lifelong adaptation that does not depend on continuous labeled supervision. In this perspective, a practical deployment pathway is to use labeled data to obtain an accurate initialization, as done in this work, followed by low-overhead online adaptation through local, hardware-native learning rules. In particular, three-factor learning with eligibility traces \cite{gerstner2018eligibility, fremaux2016neuromodulated} offers a hardware-compatible alternative to backpropagation through time by combining local synaptic traces with a global modulatory signal. Ferroelectric synapses have also demonstrated plasticity behaviors that can support Hebbian learning \cite{garg2026unsupervised}, suggesting a feasible route toward device-native learning in future systems.

\section{Conclusion}

This work demonstrates that programmable memristive hardware can support adaptive spiking neural networks for EEG-based brain-computer interfaces, enabling learning directly within a memristive synapse model. Network accuracy is largely preserved when transitioning from an ideal digital implementation to a memristive realization with nonlinear and asymmetric update dynamics, indicating that the proposed learning framework is tolerant to key device-level non-idealities. These results support the practical feasibility of deploying spiking neural networks on memristive neuromorphic hardware. Our results further highlight that such SNNs are robust to limited weight resolution and programming noise, as evidenced by the quantization and additive noise experiments. Importantly, we demonstrate that residual performance degradation can be recovered with only a few on-device learning epochs, mitigating the need for extensive reprogramming and thereby contributing to improved device lifetime. 

From a BCI perspective, the ability to perform subject-specific adaptation directly on hardware is particularly significant, as it addresses the inherent non-stationarity and inter-subject variability of EEG signals. Rather than relying on fully subject-agnostic models or repeated offline retraining, programmable memristive synapses enable post-deployment personalization with limited energy and endurance overhead. This capability is critical for wearable and implantable BCI systems intended for long-term, real-world operation. The deployment strategies and learning mechanisms developed here are not limited to motor-imagery decoding, but are broadly relevant to other neural and wearable bio-signal processing tasks. Future work will therefore focus on scaling to larger networks and full on-hardware implementations, further bridging the gap between brain-inspired algorithms and practical, personalized neuromorphic hardware systems.

\section*{Data Availability}

The PhysioNet EEG dataset used in this study is publicly available at:\\ https://physionet.org/content/eegmmidb/1.0.0/. \\
\newline
The experimental data displayed in Fig. 2 and Fig. 3 is publicly available at:\\ https://doi.org/10.5281/zenodo.19928453

\section*{Code Availability}

The code used in this study is publicly available at https://github.com/NEO-ETHZ/EEG-Ferro.

\section*{Acknowledgments}
We thank the Binning and Rohrer Nanotechnology Center, in particular U.\ Drechsler and M. Stiefel. 

\section*{Funding}
Research funded by Swiss National Foundation for Science under project ROSUBIO \#218438, by SERI under the initiative SwissChips, and by Horizon Europe under the Chips JU project ViTFOX \#101194368.

\section*{Competing Interests Statement}
The authors declare no competing interests.



\printbibliography

\end{document}